\documentclass[a4paper,fleqn]{cas-dc}

\usepackage[numbers]{natbib}

\usepackage{amssymb}
\usepackage{algorithm}
\usepackage{algorithmic}
\usepackage{newfloat}
\usepackage{listings}
\usepackage{amsmath,amsfonts}
\usepackage{array}
\usepackage{subcaption}
\usepackage{lineno}
\usepackage{textcomp}
\usepackage{makecell}
\usepackage{stfloats} 
\usepackage{url}
\usepackage{diagbox}
\usepackage{verbatim}
\usepackage{graphicx}
\usepackage{bbding}
\usepackage{pifont}
\usepackage{setspace} 
\usepackage{booktabs}
\usepackage{multirow}
\usepackage{balance} 
\usepackage{float}
\usepackage{bbding}
\usepackage[dvipsnames]{xcolor}
\usepackage{color}
\usepackage{colortbl}

\def\tsc#1{\csdef{#1}{\textsc{\lowercase{#1}}\xspace}}
\tsc{WGM}
\tsc{QE}

\begin{document}
\let\WriteBookmarks\relax
\def\floatpagepagefraction{1}
\def\textpagefraction{.001}

\shorttitle{}    

\shortauthors{}  

\title [mode = title]{Task-Specific Prompt with Global Context for Multi-Task Graph Pre-Training}  



%

\author[1]{Zhiyang Qiu}[orcid=0009-0005-1149-6472]


\ead{qzhiyang@e.gzhu.edu.cn}



\author[1]{Yangtao Wang}[orcid=0000-0003-4605-9270]

\cormark[1]
\cortext[1]{Corresponding author}

\ead{ytaowang@gzhu.edu.cn}

\author[2]{Xiaocui Li}[orcid=0000-0002-5971-2331]


\ead{Xiaocuiworld@163.com}

\author[1]{Yanzhao Xie}[orcid=0000-0002-9274-2807]


\ead{yzhx@gzhu.edu.cn}


\author[1]{Siyuan Chen}[orcid=0000-0001-9272-4804]


\ead{chensiyuan@gzhu.edu.cn}

\author[1]{Wensheng Zhang}[orcid=0000-0003-0752-941X]


\ead{zhangwenshengia@hotmail.com}

\affiliation[1]{organization={School of Computer Science and Cyber Engineering, Guangzhou University},
            city={Guang Zhou},
            postcode={510006}, 
            state={Guangdong},
            country={China}}

\affiliation[2]{organization={Hunan University of Technology and Business},
            city={Changsha},
            postcode={410205}, 
            state={Hunan},
            country={China}}




\begin{abstract}
Graph prompt learning is an effective paradigm to adapt pre-trained graph models to downstream tasks in low-resource scenarios. However, existing multi-task graph pre-training frameworks generally use randomly initialized prompts, leading to poor alignment between the prompt space, pretext objectives and graph structural characteristics. This greatly weakens the task relevance, structural awareness and transferability of prompt representations. To address this challenge, we propose TPGC, a dual-prior prompt initialization solution that explicitly models the synergy between task prior and structural prior. Specifically, the Task-Prior Injection Module first conducts a short homologous multi-task pre-training on an auxiliary graph, enabling prompt initialization to inherit optimization preferences associated with multiple pretext tasks. Built on the task-aware representations, the Structure-Prior Injection Module further extracts transferable global structural context from the auxiliary graph, converting it into layer-wise prompt vectors by aggregating structurally informative node embeddings. Extensive experiments on 6 mainstream benchmarks covering node and graph classification show that TPGC achieves consistently better performance under few-shot settings than state-of-the-art baselines, with fewer downstream tunable parameters and lower runtime. The code is available at \textcolor{blue}{\url{https://github.com/Virgilqiu/TPGC}}.
\end{abstract}


\maketitle

\section{Introduction}
\label{sec:Introduction}
Graph neural networks (GNNs) have become a fundamental paradigm for learning on graph-structured data by jointly modeling node attributes and relational dependencies~\cite{kipf2017semi,velickovic2018graph}. By extending neural representation learning to irregular graph domains, GNNs have achieved remarkable achievements in recommendation, molecular property prediction, biological network analysis, and knowledge graph reasoning~\cite{hamilton2017inductive,wu2021comprehensive}. However, most existing GNN-based solutions rely heavily on large amounts of task-specific labeled data, which becomes impractical and encounters severe performance bottlenecks in low-resource graph domains. To alleviate this issue, graph pre-training has emerged as an important research direction, aiming to learn transferable graph knowledge from large-scale unlabeled graphs~\cite{hu2020strategies,hou2022graphmae,hu2020gptgnn}. Recent contrastive and mutual-information-based pre-training methods further improve downstream adaptation in few-shot scenarios by designing effective self-supervised objectives~\cite{you2020graphcl,velickovic2019dgi,qiu2020gcc}.

Existing graph pre-training methods like DGI~\cite{velickovic2019dgi} and GraphCL~\cite{you2020graphcl} generally follow a common pre-train-then-transfer paradigm, where a graph encoder is first optimized on auxiliary or unlabeled graphs and then adapted to downstream tasks such as node classification, graph classification, and link prediction. Building upon this line of research, graph prompt learning has recently emerged as a promising lightweight paradigm, which bridges graph pre-training and downstream adaptation by introducing learnable prompts into the input or hidden representation space. Representative studies (such as GPPT~\cite{sun2022gppt}, GraphPrompt~\cite{liu2023graphprompt}, GPF~\cite{fang2023gpf}, MultiGPrompt~\cite{yu2024multigprompt}, ProNoG~\cite{pronog2024nonhomophilic}, and MKGPL~\cite{DBLP:journals/pr/XieQWCFTZ26}) have consistently demonstrated the effectiveness of prompt-based adaptation across various graph tasks. Despite these advancements, the prompt information used during the graph pre-training phase is randomly initialized, making it irrelevant to downstream tasks. In contrast, the importance of prompt initialization has already attracted widespread attention in numerous fields. For instance, Lester et al.~\cite{lester2021power} point out that prompt tuning is sensitive to initialization: semantically informed initialization is often more effective than purely random initialization. Wu et al.~\cite{wu2023infoprompt} further reveal that soft prompt tuning is highly sensitive to prompt initialization and seek more task-informative prompt initialization from an information-theoretic perspective. IAPT~\cite{zhu2024iapt} generates instance-aware soft prompts for each input instruction, showing that prompt representations should be conditioned on task or input semantics rather than treated as fixed random tokens. MVLPT~\cite{shen2024mvlpt} learns transferable prompts from multiple source vision tasks to initialize target-task prompts, demonstrating that cross-task prompt knowledge can improve few-shot vision-language adaptation. These findings indicate that prompt initialization is not a trivial implementation detail, but a key factor affecting optimization stability and task adaptation. Nevertheless, this insight has not yet been systematically introduced into graph pre-training with prompt learning.

\begin{figure*}[t]
    \centering
    \includegraphics[width=0.98\textwidth]{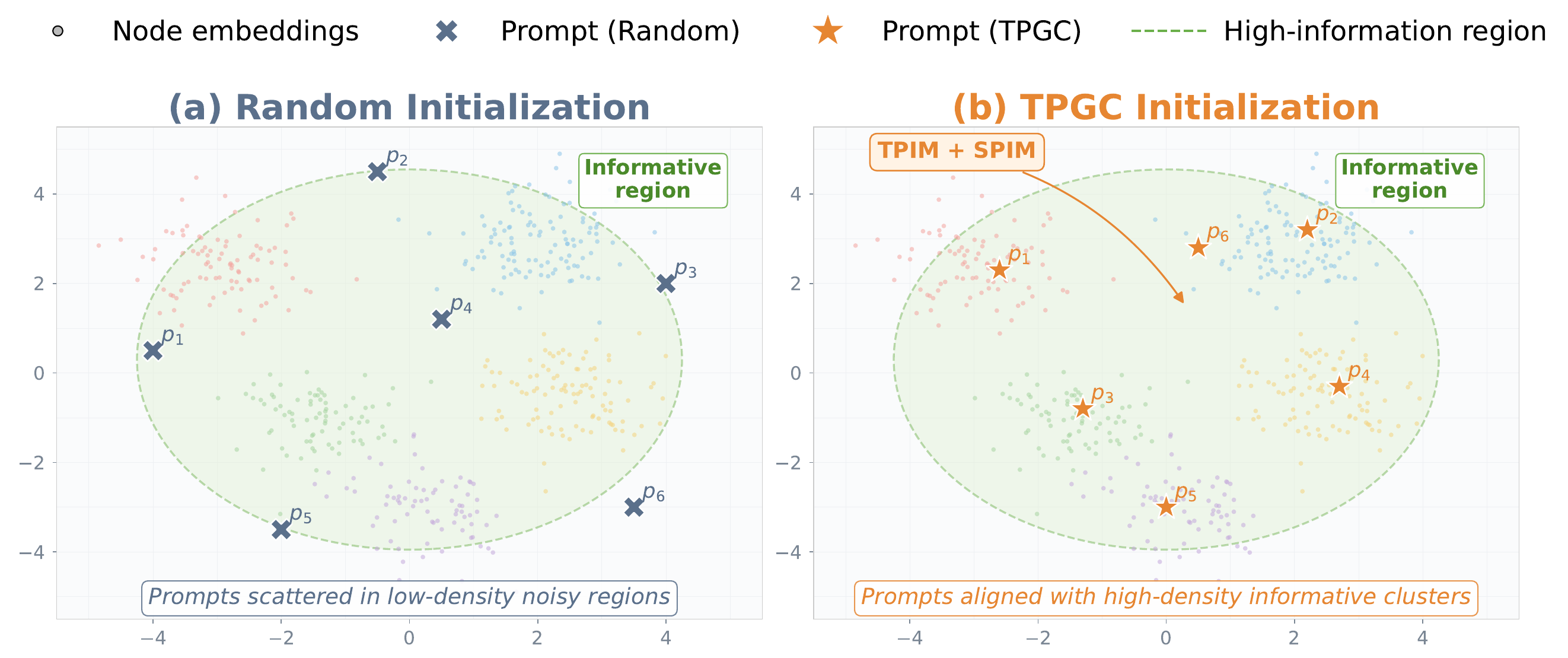}
    \caption{Visualization of prompt initialization quality in the representation space. (a): The left illustrates that random initialization places prompt vectors in arbitrary low-density regions, far from informative node clusters. (b): The right shows that TPGC initialization aligns prompt vectors with high-density and semantically meaningful regions of the node embedding space, providing a more informative and structurally aligned starting point for subsequent pre-training.}
    \label{initialization_quality}
\end{figure*}

Based on the aforementioned analysis and observations, the existing graph prompt learning paradigm still leaves two key challenges insufficiently addressed. \textbf{(1) How to endow prompt initialization with task-relevant information.} Current methods usually initialize pre-training prompts randomly, without providing an explicit mechanism to encode information associated with pre-training objectives at the initialization stage. As a result, the initialized prompts are often irrelevant to the target pretext tasks and cannot serve as effective task-aware starting points before target-domain pre-training. \textbf{(2) How to endow prompt initialization with graph structural information.} Unlike Euclidean data, the semantics of graph data are determined not only by node attributes, but also more fundamentally by topological distributions, neighborhood relations, and global structural patterns. Existing random prompt initialization in graph prompt learning does not exploit such intrinsic structural information of graph data to guide prompt construction. This issue can be intuitively understood from Figure~\ref{initialization_quality}(a), where randomly initialized prompts tend to fall into arbitrary low-density noisy regions of the representation space, far away from meaningful node clusters. Consequently, the initialized prompts fail to offer sufficiently discriminative structural support for different tasks, and can be easily biased toward a specific task during optimization, thus undermining the effectiveness of multi-task collaboration. As a result, how to explicitly inject \textbf{task-specific prior} and make prompt initialization \textbf{graph-structure-aware} remains a core challenge in multi-task graph pre-training.

To address the above challenges, we propose \textbf{T}ask-Specific \textbf{P}rompt with \textbf{G}lobal \textbf{C}ontext for Multi-Task Graph Pre-Training (termed TPGC), which aims to construct prompts that are both task-specific and global-context-aware, thereby providing a more informative prompt initialization to boost graph pre-training. Specifically, TPGC decomposes prompt initialization into two collaborative components: a task-prior injection module (TPIM) and a structure-prior injection module (SPIM). (1) TPIM aims to inject task-specific prior associated with pre-training objectives into prompts before target-domain pre-training, thereby alleviating the mismatch between randomly initialized prompts and pre-training objectives. Notably, this task-prior extraction process is highly efficient: on an auxiliary graph that is highly related to the target graph in terms of domain semantics and structural characteristics, only one epoch of homologous pre-training is sufficient to provide an effective task-related prior for prompt initialization. 
(2) Furthermore, SPIM captures transferable structural prior from the global context of the auxiliary graph and injects it into the prompt space, thereby making the initialized prompts more aligned with informative regions in the graph representation space. This effect is conceptually illustrated by Figure~\ref{initialization_quality}(b), where the initialized prompts are no longer scattered in noisy areas, but instead lie close to high-density and semantically meaningful node clusters. In this way, TPIM mainly improves the task relevance and optimization stability of prompt initialization, while SPIM further enhances its structural awareness and cross-graph transferability. Notably, the prompt initialization stages of SPIM and TPIM can be completed offline on the auxiliary graph, so the proposed method does not incur noticeable additional computation in the target-domain pre-training stage. The prompts initialized by these two modules are then used in subsequent multi-task pre-training and transferred to downstream prompt tuning, enabling lightweight graph adaptation without modifying the backbone architecture.

The main contributions of this study are summarized as follows:
\begin{itemize}
    \item \textbf{New perspective.} To the best of our knowledge, this study is the first to revisit multi-task graph pre-training from the perspective of prompt initialization rather than prompt design/tuning alone. Different from conventional random initialization, the proposed initialization paradigm explicitly constructs task-specific prompts with global context, so that prompt initialization can simultaneously preserve task-specific prior and transferable structural knowledge.
    \item \textbf{New method.} We propose a novel prompt initialization method, termed TPGC, which constructs prompts through two complementary components. Specifically, TPIM extracts task-specific prior through only one epoch of homologous pre-training on an auxiliary graph, while SPIM captures transferable structural prior from the global context of the auxiliary graph and injects it into the prompt space. In this way, TPGC produces prompts that are both task-specific and global-context-aware for multi-task graph pre-training.
    \item \textbf{High performance.} We conduct extensive experiments on multiple mainstream node-level and graph-level benchmark datasets. Experimental results show that our TPGC consistently outperforms strong graph prompt learning baselines under various few-shot settings. In addition, TPGC demonstrates strong robustness across all few-shot settings, while introducing lower downstream tunable parameter count and runtime.
\end{itemize}

\section{Related Works}
\label{sec:Related Works}
 
\subsection{Graph Pre-Training}

Graph pre-training~\cite{DBLP:journals/tkde/YuLFLCZ24,DBLP:conf/aaai/HoangL25,DBLP:journals/tdsc/XuZWLHF26,DBLP:journals/pr/HuangZJJLM26} seeks to distill transferable knowledge from unlabeled graph data, thereby reducing the reliance of graph learning on expensive task-specific annotations. Following the success of self-supervised learning, a growing body of work has shown that graph encoders can be effectively improved by optimizing carefully designed pretext tasks without manual labels \cite{hu2020strategies,hu2020gptgnn}. These pretext tasks are usually constructed to capture intrinsic graph signals from different views, including structural dependency, attribute semantics, local-global consistency, and topology-aware context \cite{hu2020strategies,hou2022graphmae,xia2023molebert}. Along this line, some studies improve graph representation learning through bootstrap-based or contrastive-style objectives that enhance invariance and representation robustness, while others adopt generative or reconstruction-based objectives to recover informative graph patterns and preserve semantic content \cite{hou2022graphmae,hu2020gptgnn,thakoor2022bgrl,hou2023graphmae2}. However, pretext tasks defined from a single perspective often provide only partial supervision for complex graph data, which has motivated recent efforts to incorporate multiple pretext tasks into unified pre-training frameworks \cite{hu2020strategies,xia2023molebert}. 

When multiple tasks are optimized jointly, the model may still suffer from task interference, since different objectives can emphasize different graph properties and lead to inconsistent optimization directions. In addition, recent studies have further extended graph pre-training to broader transfer settings, where transferable knowledge is expected to generalize across heterogeneous graph domains rather than a single distribution \cite{lin2025unified}. Nevertheless, a central challenge remains how to effectively coordinate diverse pretext tasks while improving the alignment between transferred knowledge and downstream objectives. Different from existing studies that mainly refine pretext objectives or encoder architectures, our method enhances multi-task graph pre-training by introducing task-specific prompts with global contextual guidance, enabling more effective integration of transferable task semantics and structural information.
  
\subsection{Graph Prompt Learning}

Prompt learning~\cite{DBLP:conf/aaai/WuWZZLLH23,DBLP:journals/ijcv/XuZSCLCW25,DBLP:journals/pr/DengWXLTFZ26} was first popularized in vision and language research as a parameter-efficient paradigm for adapting pre-trained models to downstream tasks. Instead of updating all model parameters, prompt-based methods introduce learnable context vectors to steer model behavior, preserving general knowledge while improving task adaptation~\cite{zhou2022coop,zhou2022cocoop,khattak2023maple}. Inspired by these advances, prompt learning has been introduced into graph representation learning to bridge pre-training and downstream adaptation. Early studies reformulated graph tasks into prompt-based forms for lightweight adaptation~\cite{sun2022gppt,liu2023graphprompt}, while subsequent work improved generality by unifying multiple tasks under shared prompting frameworks~\cite{sun2023allinone,fang2023gpf}. More recently, graph prompt learning~\cite{DBLP:conf/icml/AiLZ25,DBLP:journals/tkde/YangHZS26,DBLP:journals/pr/XieQWCFTZ26} has extended to multi-task pre-training, heterogeneous graph learning, and fairness-aware modeling~\cite{yu2024multigprompt,sun2025generalizable,jiao2025hgmp,li2025fairness}. Despite these advances, existing methods mainly emphasize prompt design or downstream adaptation, while paying less attention to how prompts can effectively capture transferable task semantics and graph structural cues when multiple pre-training objectives coexist.

Prompt initialization has received increasing attention in natural language processing, revealing that prompt tuning is highly sensitive to initialization and semantically informed initialization is often more effective than random initialization~\cite{lester2021power,wu2023infoprompt}. TPV shows that task prompt vectors can effectively initialize prompt tuning on related tasks through multi-task soft-prompt transfer~\cite{belanec2025tpv}. VPTTA~\cite{chen2024vptta} also demonstrates that prompt initialization can be enhanced with memory-based visual knowledge to support more reliable test-time adaptation. These findings suggest that prompt initialization is a key factor affecting the effectiveness of prompt-based learning. In graph prompt learning, however, initialization remains insufficiently explored. Although meta-learning-based initialization has been proposed in a unified prompting framework~\cite{sun2023allinone}, such designs mainly target single pretext-task settings and do not address multi-task graph pre-training, where multiple tasks jointly optimize a shared prompt space. To address this limitation, our method studies prompt initialization from the perspective of multi-task graph pre-training by incorporating task-specific prior and global structural context into the initialization process. In this way, we provide a more informative starting point for shared prompts, enabling better multi-task coordination and stronger transfer of structural and semantic knowledge.

\section{Proposed Methodology}
\label{sec:method}
In this section, we present and formulate the detailed workflow of our designed TPGC, i.e., an innovative prompt initialization strategy for multi-task graph pre-training. By jointly injecting task-aware prior and transferable global structural context into the prompt space, TPGC produces a more informative initialization that improves the alignment between pretext objectives and graph representations.

\begin{figure*}[t]
    \centering
    \includegraphics[width=1\linewidth]{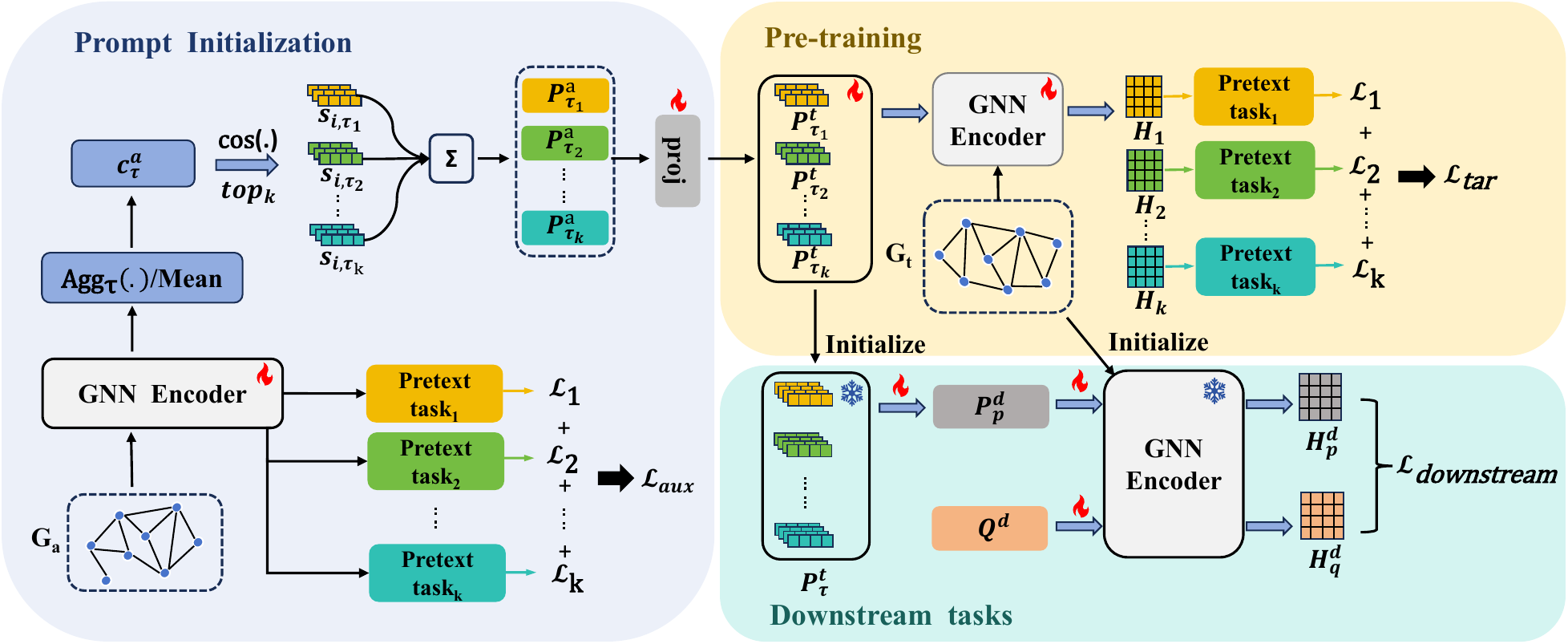}
    \caption{The overall architecture of our proposed TPGC consists of three successive stages, namely prompt initialization, target-domain multi-task pre-training, and downstream prompt-based adaptation. Note that the "fire" means the corresponding modules/parameters are learnable, while the "snow" means those are frozen.}
    \label{figure:overall architecture}
\end{figure*}

\subsection{Overall Architecture}
As illustrated in Figure~\ref{figure:overall architecture}, TPGC consists of three successive stages, namely prompt initialization, target-domain multi-task pre-training, and downstream prompt-based adaptation. (1) In the first stage, instead of initializing prompts from random noise, we construct task-specific prompt groups through a dual-prior injection process. Specifically, TPIM first performs a short auxiliary multi-task pre-training process on a homologous auxiliary graph $G^a$, so that the shared graph encoder can absorb optimization  preference associated with different pretext objectives and evolve into a task-aware auxiliary encoder. Based on this optimized encoder, SPIM further extracts transferable global structural context from auxiliary graph representations, computes task-relevant similarity scores, and aggregates informative node embeddings into layer-wise initialized prompt vectors. After an input-layer projection for cross-graph feature alignment, the resulting initialized prompt groups are transferred to the target graph as the starting prompts for subsequent optimization.
(2) In the second stage, the initialized prompt groups are injected into the target graph encoder and jointly optimized with the shared encoder under multiple pretext objectives, yielding a target-domain pre-trained encoder together with task-aware prompt knowledge. (3) In the final stage, the pre-trained prompt groups are frozen and reused as transferable prompt prior for downstream adaptation, while another downstream-task-specific prompt group is introduced to complement task-specific flexibility. The two prompt branches are then fused to support downstream prediction in a lightweight manner. Through this three-stage pipeline, TPGC establishes a coherent flow from informative prompt initialization to target-domain pre-training and finally to downstream prompt adaptation.

\subsection{Prompt Initialization}
Before target-domain multi-task pre-training, we introduce a dedicated prompt initialization stage to construct a more informative starting point for each task-specific prompt group. Rather than directly optimizing prompts from unconstrained initial values, we first organize the initialization process into two successive modules that jointly determine where the prompt starts in the representation space. Specifically, TPIM performs a short auxiliary multi-task pre-training step on a homologous auxiliary graph, so that the shared graph encoder can first absorb optimization preference associated with the pretext objectives and evolve into a task-aware auxiliary encoder. Built upon this optimized encoder, SPIM further extracts transferable global structural context from auxiliary graph representations and converts it into layer-wise prompt vectors by selecting and aggregating structurally informative node embeddings. In this way, TPIM provides the task-related basis for prompt construction, while SPIM further refines this basis with transferable structural guidance, and their combination finally yields the initialized prompt groups that will be transferred to the target graph for subsequent multi-task pre-training.

\subsubsection{Task-Prior Injection Module (TPIM)}
In the graph setting, different encoder layers may emphasize different semantic levels of representation. The input layer mainly preserves raw node attributes and is therefore more relevant to node-level pretext objectives, whereas the hidden and output layers gradually encode higher-order structural context and are thus more beneficial for relation-oriented or graph-level objectives. Inspired by prompt-based graph pre-training frameworks, we assign layer-wise pretext tokens to every pretext task so that each task can inject its own inductive preference into the input, hidden, and output layers of the encoder in a unified manner.

Specifically, given a graph $G$, a graph encoder with $L$ layers, and $K$ pretext tasks $\mathcal{S} = \{\tau_1, \tau_2, \dots, \tau_K\}$, for each pretext task $\tau_k \in \mathcal{S}$, we associate a layer-wise prompt group $P_{\tau_k} = \{p_{\tau_k}^{(0)}, p_{\tau_k}^{(1)}, \dots, p_{\tau_k}^{(L)}\}$ with the encoder, where $p_{\tau_k}^{(l)}$ denotes the prompt token injected at the $l$-th encoder layer for the $k$-th pretext task. In this way, each task is equipped with its own prompt group spanning all encoder layers, enabling task-specific modulation of graph representations while preserving a shared backbone for multi-task optimization.

To provide transferable prior before target-domain optimization, we introduce an auxiliary graph that is semantically and structurally related to the target graph. For example, when Cora is used as the target graph, Citeseer can serve as the auxiliary graph because both belong to citation-network domains and exhibit similar semantic and structural properties. As shown in Figure~\ref{figure:overall architecture}, let the auxiliary graph and the target graph be denoted by $G^a = (V^a, E^a, X^a, A^a)$ and $G^t = (V^t, E^t, X^t, A^t)$, respectively, where $X$ and $A$ denote the node feature matrix and adjacency matrix. Based on the above pretext-task formulation, our goal here is to construct an informative initialized prompt group $P_{\tau}^{t}$ for each task $\tau \in \mathcal{S}$ before target-domain optimization starts.

Then, we first perform a short homologous pre-training process on an auxiliary graph $G^a$ that is semantically and structurally related to the target graph $G^t$. This auxiliary optimization follows the same pretext-task setting as the target stage, so that the shared graph encoder can absorb supervisory signals jointly induced by multiple pretext tasks. To express this process more explicitly, let $\mathcal{S}$ denote the set of pretext tasks considered during auxiliary optimization. The auxiliary multi-task pre-training objective is written as:

\begin{equation}
\label{eq:aux_objective}
\mathcal{L}_{\mathrm{aux}}(f_{\theta} \rightarrow f_{\theta^{a}}) = \sum_{\tau \in \mathcal{S}} \lambda_{\tau} \mathcal{L}_{\tau}^{a},
\end{equation}
where $\mathcal{L}_{\tau}^{a}$ denotes the auxiliary loss of pretext task $\tau$ on $G^a$, and $\lambda_{\tau}$ is the corresponding task weight. After the auxiliary multi-task pre-training stage in TPIM, the graph encoder is updated from the initial encoder \(f_{\theta}\) to the task-aware auxiliary encoder \(f_{\theta^{a}}\), which is then used by SPIM to extract transferable structural context for prompt initialization.

\subsubsection{Structure-Prior Injection Module (SPIM)}
While TPIM provides a task-aware encoder, it does not explicitly determine how the prompt vectors should be initialized in the graph representation space. To address this issue, we further propose a Structure-Prior Injection Module (SPIM), which uses the optimized encoder parameters $\theta^{a}$ to extract transferable global structural context and convert it into layer-wise prompt initialization. Specifically, for each layer $l \in \{0,1,\dots,L\}$, we first obtain the corresponding auxiliary node representations under the condition of $\theta^{a}$:
\begin{equation}
\label{eq:layerwise_representation}
H^{a} = f_{\theta^{a}}(X^a, A^a),
\end{equation}
where $H^{a,(l)}$ denotes the feature matrix at the $l$-th encoder layer on the auxiliary graph.  Based on these layer-wise node representations, we construct a task-related global embedding $c_{\tau}^{a,(l)}$ for each pretext task and prompting location:
\begin{equation}
\label{eq:global_embedding}
c_{\tau}^{a,(l)} =
\begin{cases}
\mathrm{Agg}_{\tau}(H^{a,(l)}) \\
\mathrm{Mean}(H^{a,(l)})
\end{cases}.
\end{equation}
Note that for pretext tasks such as DGI~\cite{velickovic2019dgi} and GraphCL~\cite{you2020graphcl}, which emphasize local-global consistency or augmentation-invariant discrimination, $c_{\tau}^{a,(l)}$ is obtained by applying a task-aware aggregation operator $\mathrm{Agg}_{\tau}(\cdot)$ to $H^{a,(l)}$ so as to preserve informative global contextual cues. By contrast, for relation-oriented tasks such as link prediction and edge reconstruction, $c_{\tau}^{a,(l)}$ is computed by global mean pooling over $H^{a,(l)}$, since these tasks rely more on stable pairwise regularities and overall connectivity statistics.
After obtaining $c_{\tau}^{a,(l)}$, we compute the cosine similarity between each node embedding and the corresponding task-related global embedding:   
\begin{equation}
\label{eq:similarity_score}
s_{i,\tau}^{(l)} = \mathrm{cos}(h_i^{a,(l)}, c_{\tau}^{a,(l)}).
\end{equation}

We then select the top-$k$ nodes with the highest similarity scores and denote the selected node set by $\mathcal{S}_{\tau}^{(l)}$. Their embeddings are aggregated with similarity-aware weights to obtain the initialized prompt vector for pretext task $\tau$ at layer $l$:
\begin{equation}
\label{eq:aux_prompt_initialization}
{p}_{\tau}^{a,(l)} = \mathrm{Norm}\left(\sum_{i \in \mathcal{S}_{\tau}^{(l)}} s_{i,\tau}^{(l)} \, h_i^{a,(l)} \right),
\end{equation}
where $\mathrm{Norm}(\cdot)$ denotes feature normalization. This operation directly determines the initialized prompt parameter for task $\tau$ at layer $l$. Repeating the same procedure over all prompting layers yields the layer-wise initialized prompts for the corresponding pretext task.

Since pre-training also introduces prompt parameters at the input layer, an additional dimension adaptation step is required when the layer-wise prompt group is transferred across graphs. When the auxiliary graph and the target graph have different input feature dimensions, the input-layer prompt is mapped from the auxiliary feature space to the target feature space through a learnable projection matrix $W_{\mathrm{proj}} \in \mathbb{R}^{d_t \times d_a}$, yielding $p_{\tau}^{t,(0)}$ from $p_{\tau}^{a,(0)}$. For hidden prompting layers, the initialized prompts are directly transferred, i.e., $p_{\tau}^{t,(l)}$ is directly obtained from $p_{\tau}^{a,(l)}$ for $l = 1, 2, \dots, L$, since their dimensions have already been unified by the graph encoder. Accordingly, for each pretext task $\tau_k \in \mathcal{S}$, the final initialized prompt group on the target graph is given by:
\begin{equation}
\label{eq:target_prompt_group}
P_{\tau_k}^{t} = \{p_{\tau_k}^{t,(0)}, p_{\tau_k}^{t,(1)}, \dots, p_{\tau_k}^{t,(L)}\}.
\end{equation}

After the above dual-prior injection process, the prompt group no longer starts from a random distribution, but from an initialization that jointly encodes task-aware optimization preference and transferable structural context. This initialized prompt group is then used as the starting point of multi-task pre-training on the target graph $G^t$. 

\subsection{Target-Domain Multi-Task Pre-Training}
After the above dual-prior injection process, each pretext task $\tau_k \in \mathcal{S}$ is equipped with an initialized prompt group $P_{\tau_k}^{t}$ on the target graph. These initialized prompt tokens are then injected into the shared graph encoder and used to guide target-domain multi-task pre-training. In this way, the target-stage optimization starts from transferred task-aware and structure-aware prompt initialization rather than from random prompt parameters.

Starting from the transferred auxiliary encoder $f_{\theta^{a}}$, the target graph is encoded into layer-wise node representations $H^{t}$, where $H^{t,(l)}$ denotes the feature matrix at the $l$-th encoder layer. For the $l$-th layer, the corresponding prompt token rewrites the current feature matrix by row-wise modulation, and the modulated output is further propagated to the next layer. When $l$ < $L$, the next layer will be generated as:
\begin{equation}
\label{eq:target_prompt_modulation}
H^{t,(l+1)} = MessagePassing(p^{t,(l)} \odot H^{t,(l)},A^t,f_{\theta^{a}}),
\end{equation}
where $\odot$ denotes row-wise element-wise multiplication between the prompt token and the layer-wise feature matrix. Repeating this process over all layers yields a sequence of prompt-conditioned representations for the $k$-th pretext task, denoted by $\{H_{\tau_k}^{t,(l)}\}_{l=0}^{L}$. These layer-wise representations are then aggregated into the final task-specific representation:
\begin{equation}
\label{eq:target_task_representation}
H^{t}_{\tau_k} = \sum_{l=0}^{L} \alpha_l H_{\tau_k}^{t,(l)},
\end{equation}
where $\alpha_l$ denotes the fusion weight of the $l$-th encoder layer. This weighted aggregation allows the final pre-training representation to preserve both shallow feature information and deep structural semantics, so that each pretext task can make fuller use of multi-level graph representations.

Based on the resulting task-specific representation, the target-domain multi-task pre-training objective is defined as:
\begin{equation}
\label{eq:target_objective}
\mathcal{L}_{\mathrm{tar}}(H^{t}_{\tau_k},f_{\theta^{a}} \rightarrow f_{\theta^{t}}) = \sum_{\tau \in \mathcal{S}} \lambda_{\tau} \mathcal{L}_{\tau}^{t}.
\end{equation}
Under this objective, the initialized prompt tokens and the shared encoder are jointly optimized on the target graph. Starting from the transferred auxiliary encoder \(f_{\theta^{a}}\), the target-domain multi-task pre-training stage further updates the graph encoder to \(f_{\theta^{t}}\) under the guidance of the initialized prompt groups. Therefore, the target-stage pre-training inherits both the task prior encoded by TPIM and the structure prior injected by SPIM, while preserving the original multi-task collaborative learning paradigm.

\subsection{Prompt-Based Downstream Adaptation}
After target-domain multi-task pre-training, the graph encoder has been updated from $f_{\theta^{a}}$ to $f_{\theta^{t}}$. The downstream stage is built on $f_{\theta^{t}}$, where the transferred prompt groups $\{P_{\tau_k}^{t}\}_{k=1}^{K}$ are frozen as a prompt prior, while another randomly initialized prompt group is introduced for downstream adaptation.

Specifically, for each layer $l \in \{0,1,\dots,L\}$, we aggregate the frozen prompts transferred from all pretext tasks with learnable weights to obtain a layer-wise pre-trained prompt for downstream task $d$:
\begin{equation}
\label{eq:downstream_pretrained_prompt}
p_{\mathrm{p}}^{d,(l)}=\sum_{k=1}^{K}\beta_k^{(l)}\,p_{\tau_k}^{t,(l)},
\end{equation}
where $\beta_k^{(l)}$ denotes the learnable contribution weight of the $k$-th pretext task at the $l$-th layer. Collecting the layer-wise prompts $p_{\mathrm{p}}^{d,(l)}$ over all encoder layers yields the transferred pre-trained prompt group $P_{\mathrm{p}}^{d}$. In parallel, the downstream-task-specific prompt group is defined as $Q^{d}=\{q^{d,(0)},q^{d,(1)},\\ \dots,q^{d,(L)}\}$.

To remain consistent with target-domain pre-training, downstream adaptation is also performed in a layer-wise manner. Let $E^{d,(l)}$ denote the downstream instance representations extracted from the $l$-th layer of $f_{\theta^{t}}$, where each instance corresponds to a node representation or a graph representation after readout. Based on the transferred pre-trained prompt group and the downstream-task-specific prompt group, we obtain two branch representations by weighted aggregation across layers:
\begin{equation}
\begin{aligned}
\label{eq:downstream_branch_representation}
H_{\mathrm{p}}^{d}=\sum_{l=0}^{L}\alpha_l\big(p_{\mathrm{p}}^{d,(l)}\odot E^{d,(l)}\big), \\
H_{\mathrm{q}}^{d}=\sum_{l=0}^{L}\alpha_l\big(q^{d,(l)}\odot E^{d,(l)}\big),
\end{aligned}
\end{equation}
where $\alpha_l$ denotes the learnable fusion weight of the $l$-th layer, and $\odot$ denotes row-wise element-wise modulation. The two branch representations are then fused by a learnable scalar parameter:
\begin{equation}
\label{eq:downstream_fused_representation}
H^{d}=\gamma H_{\mathrm{p}}^{d}+(1-\gamma)H_{\mathrm{q}}^{d}.
\end{equation}
where $\gamma$ is a learnable scalar balancing the two branches. Let $H^{d}=\{h_m^{d}\}$ denote the final downstream instance representations. Based on the labeled support instances, we construct a prototype $r_c^{d}$ for each class $c$ by averaging the fused representations $h_m^{d}$ belonging to that class, where $\mathcal{I}_c$ denotes the set of labeled support instances from class $c$. The downstream adaptation is then optimized under the corresponding task-specific objective $\mathcal{L}_{\mathrm{downstream}}$, which is defined over the fused instance representations and the induced class prototypes. In this way, both node-level and graph-level downstream tasks share the same transfer mechanism, where frozen pre-trained prompts provide transferable prior and downstream-task-specific prompts supply task-adaptive flexibility.

\section{Experiments}
\label{sec:experiments}
\subsection{Experimental Settings}

\subsubsection{Datasets}
We evaluate the proposed method on six widely used benchmark datasets, including two citation network datasets (\textit{Cora} \cite{mccallum2000automating} and \textit{Citeseer} \cite{giles1998citeseer}) and four bioinformatics/chemical graph datasets (\textit{PROTEINS} \cite{dobson2003distinguishing}, \textit{ENZYMES} \cite{schomburg2004brenda}, \textit{COX2} \cite{sutherland2003spline}, and \textit{BZR} \cite{morris2020tudataset}). These datasets cover both node-level and graph-level prediction settings, which helps comprehensively verify the effectiveness of our method across different graph tasks.

\begin{itemize}
\item \textbf{Cora} is a widely used citation network dataset for node classification \cite{mccallum2000automating}. It contains 2,708 scientific publications connected by citation links, where each node denotes a paper, each edge denotes a citation relation, and each paper is represented by a 1,433-dimensional bag-of-words feature vector. With both node attributes and graph topology available, Cora is a standard benchmark for evaluating node-level representation learning and prompt-based adaptation across 7 research categories.

\item \textbf{Citeseer} is another classic citation network benchmark for node classification \cite{giles1998citeseer}. It consists of 3,327 scientific publications and 4,732 citation links, where each node corresponds to a document described by a 3,703-dimensional feature vector. Compared with Cora, Citeseer is relatively sparser and more challenging, making it suitable for evaluating the robustness and transferability of graph prompt learning methods over 6 subject categories.

\item \textbf{PROTEINS} is a protein graph dataset commonly used in graph learning \cite{dobson2003distinguishing}. It contains 1,113 protein graphs with 2 graph labels, where each graph represents a protein, nodes usually correspond to secondary structure elements, and edges describe their structural or spatial relations. Because it preserves both structural patterns and node-level biological information, PROTEINS can be used to evaluate both graph-level classification performance and node-level adaptation ability under our experimental setting.

\item \textbf{ENZYMES} is a benchmark dataset of protein tertiary structures derived from the BRENDA enzyme database \cite{schomburg2004brenda}. It contains 600 enzyme graphs, where each graph represents an enzyme structure and the task is to classify it into one of 6 enzyme commission (EC) top-level classes. Owing to its relatively fine-grained category division and graph topology that reflects interactions among structural components, ENZYMES is a representative benchmark for evaluating both node-level and graph-level graph learning methods.

\item \textbf{COX2} is a molecular graph classification dataset containing 467 molecule graphs with binary labels \cite{sutherland2003spline}. Each graph denotes a chemical compound, where nodes represent atoms and edges represent chemical bonds, and the task is to distinguish whether the compound is associated with the target biochemical property related to cyclooxygenase-2 (COX2). Since molecular graphs usually exhibit diverse local substructures, COX2 is suitable for assessing the ability of the model to capture discriminative structural patterns at the graph level.

\item \textbf{BZR} is a molecular graph benchmark from the TUDataset collection \cite{morris2020tudataset}. It contains 405 molecule graphs with 2 graph classes, where each graph corresponds to a chemical compound with atoms as nodes and bonds as edges. Due to its relatively small scale, nontrivial structural diversity, and classification target related to biochemical activity on benzodiazepine receptors, BZR is frequently used to evaluate the generalization ability of graph classification models in few-shot and low-resource settings.
\end{itemize}

In our experiments, we evaluate the proposed method on both node classification and graph classification tasks.

\begin{itemize}
\item \textbf{Node classification datasets:} \textit{Cora}, \textit{Citeseer}, \textit{PROTEINS}, and \textit{ENZYMES}. Among them, \textit{Cora} and \textit{Citeseer} are standard citation-network node classification benchmarks, while \textit{PROTEINS} and \textit{ENZYMES} are further adopted under our node-level setting to verify the transferability of the method on biological graph data.

\item \textbf{Graph classification datasets:} \textit{PROTEINS}, \textit{ENZYMES}, \textit{COX2}, and \textit{BZR}. These datasets cover protein graphs and molecular graphs, enabling us to evaluate the effectiveness of the proposed method on graph-level prediction tasks with different semantic domains and structural characteristics.
\end{itemize}

\begin{table*}[t]
\centering
\footnotesize
\caption{Performance (\%) comparisons (mean accuracy $\pm$ standard deviation) on node classification datasets under 1-shot and 5-shot settings. Note that for each dataset in each shot setting, we mark the best result in \textbf{bold}, and the second-best result in \underline{underline}.}
\label{tab:node_results}
\setlength{\tabcolsep}{6pt}
\renewcommand{\arraystretch}{1.15}
\scalebox{0.97}{
\begin{tabular}{p{0.34\textwidth}*{4}{>{\centering\arraybackslash}p{0.14\textwidth}}}
\hline
Methods & Cora & Citeseer & PROTEINS & ENZYMES \\
\hline\hline
\multicolumn{5}{c}{1-shot} \\
\hline
GCN~\cite{kipf2017semi} (ICLR 2017) & 28.57 $\pm$ 5.07 & 31.27 $\pm$ 4.53 & 43.31 $\pm$ 9.35 & 48.08 $\pm$ 4.71 \\
GAT~\cite{velickovic2018gat} (ICLR 2018) & 28.40 $\pm$ 6.25 & 30.76 $\pm$ 5.40 & 31.79 $\pm$ 20.11 & 35.32 $\pm$ 18.72 \\
\hline
DGI/InfoGraph~\cite{velickovic2019dgi} (ICLR 2019) & 54.11 $\pm$ 9.60 & 45.00 $\pm$ 9.19 & 45.22 $\pm$ 11.09 & 48.05 $\pm$ 14.83 \\
GraphCL~\cite{you2020graphcl} (NeurIPS 2020) & 51.96 $\pm$ 9.43 & 43.21 $\pm$ 9.61 & 46.15 $\pm$ 10.94 & 48.88 $\pm$ 15.98 \\
\hline
GPPT~\cite{sun2022gppt} (KDD 2022) & 15.37 $\pm$ 4.51 & 21.45 $\pm$ 3.45 & 35.15 $\pm$ 11.40 & 35.37 $\pm$ 9.37 \\
GraphPrompt~\cite{liu2023graphprompt} (WWW 2023) & 54.25 $\pm$ 9.38 & 45.34 $\pm$ 10.53 & 47.22 $\pm$ 11.05 & 53.54 $\pm$ 15.46 \\
MultiGPrompt~\cite{yu2024multigprompt} (WWW 2024) & 57.73 $\pm$ 9.96 & \underline{53.89 $\pm$ 11.68} & 48.23 $\pm$ 11.29 & 53.95 $\pm$ 15.50 \\
ProNoG~\cite{pronog2024nonhomophilic} (KDD 2025) & \underline{57.85 $\pm$ 10.18} & 48.57 $\pm$ 9.79 & \underline{48.95 $\pm$ 10.85} & \textbf{65.87 $\pm$ 21.15} \\
\hline
\rowcolor{orange!10} \textbf{TPGC (Ours)} & \textbf{58.57 $\pm$ 9.91} & \textbf{54.77 $\pm$ 11.29} & \textbf{49.10 $\pm$ 12.03} & \underline{55.45 $\pm$ 15.01} \\
\hline
\multicolumn{5}{c}{5-shot} \\
\hline
GCN~\cite{kipf2017semi} (ICLR 2017) & 52.32 $\pm$ 3.38 & 50.43 $\pm$ 2.37 & 44.23 $\pm$ 8.39 & 57.43 $\pm$ 4.27 \\
GAT~\cite{velickovic2018gat} (ICLR 2018) & 53.49 $\pm$ 3.93 & 51.06 $\pm$ 3.01 & 25.29 $\pm$ 25.34 & 50.89 $\pm$ 11.22 \\
\hline
DGI/InfoGraph~\cite{velickovic2019dgi} (ICLR 2019) & 71.31 $\pm$ 2.71 & 65.28 $\pm$ 3.37 & 47.10 $\pm$ 9.13 & 67.86 $\pm$ 12.12 \\
GraphCL~\cite{you2020graphcl} (NeurIPS 2020) & 71.41 $\pm$ 2.71 & 64.28 $\pm$ 3.37 & 48.90 $\pm$ 9.98 & 66.56 $\pm$ 11.32 \\
\hline
GPPT~\cite{sun2022gppt} (KDD 2022) & 50.46 $\pm$ 3.39 & 45.86 $\pm$ 2.72 & 26.74 $\pm$ 13.56 & 48.33 $\pm$ 7.47 \\
GraphPrompt~\cite{liu2023graphprompt} (WWW 2023) & 71.51 $\pm$ 2.71 & 65.38 $\pm$ 3.37 & 47.95 $\pm$ 11.26 & 68.26 $\pm$ 10.07 \\
MultiGPrompt~\cite{yu2024multigprompt} (WWW 2024) & \underline{72.93 $\pm$ 2.18} & \underline{68.88 $\pm$ 3.29} & 47.85 $\pm$ 9.45 & 67.99 $\pm$ 9.07 \\
ProNoG~\cite{pronog2024nonhomophilic} (KDD 2025) & 71.51 $\pm$ 2.71 & 64.59 $\pm$ 3.27 & \underline{49.08 $\pm$ 8.79} & \textbf{79.13 $\pm$ 7.06} \\
\hline
\rowcolor{orange!10} \textbf{TPGC (Ours)} & \textbf{76.13 $\pm$ 2.70} & \textbf{69.59 $\pm$ 3.17} & \textbf{49.33 $\pm$ 9.18} & \underline{69.47 $\pm$ 9.15} \\
\hline
\end{tabular}
}
\end{table*}

\subsubsection{Baselines}

To comprehensively evaluate the effectiveness of the proposed method, we compare it with three categories of representative baselines, namely end-to-end graph neural network methods, graph pre-training methods, and graph prompt learning methods.

\begin{itemize}
\item \textbf{End-to-end graph neural network} methods serve as the most direct supervised baselines, since they are trained on downstream tasks without separate pre-training or prompt tuning. GCN~\cite{kipf2017semi} performs neighborhood aggregation through graph convolutions and has become one of the most widely adopted benchmark models in graph learning. GAT~\cite{velickovic2018gat} further introduces an attention mechanism into neighborhood aggregation, enabling the model to assign adaptive importance weights to different neighbors.

\item \textbf{Graph pre-training} methods first learn transferable graph representations from unlabeled data through self-supervised objectives and then adapt the learned encoder to downstream tasks. DGI~\cite{velickovic2019dgi} maximizes the mutual information between local node embeddings and a global graph summary, thereby capturing informative structural and semantic patterns. GraphCL~\cite{you2020graphcl} learns robust graph representations by contrasting different augmented views of the same graph, and has become a representative contrastive pre-training baseline in graph representation learning.

\item \textbf{Graph prompt learning} methods aim to efficiently adapt pre-trained graph models to downstream tasks through learnable prompts. GPPT~\cite{sun2022gppt} is one of the earliest frameworks to introduce prompt tuning into graph learning, although it only supports node classification tasks. GraphPrompt~\cite{liu2023graphprompt} further unifies graph pre-training and downstream task modeling within a unified prompting framework. MultiGPrompt~\cite{yu2024multigprompt} extends this line of research to the multi-task setting by jointly modeling multiple pre-training objectives under a unified prompting paradigm. ProNoG~\cite{pronog2024nonhomophilic} combines graph pre-training with prompt learning and further considers the challenges introduced by non-homophilic graph structures. Its strong adaptability under different graph structural properties makes it a competitive recent graph prompt learning baseline in our experiments.
\end{itemize}

\begin{table*}[t]
\centering
\footnotesize
\caption{Performance (\%) comparisons (mean accuracy $\pm$ standard deviation) on graph classification datasets under 1-shot and 5-shot settings. Note that for each dataset in each shot setting, we mark the best result in \textbf{bold}, and the second-best result in \underline{underline}.}
\label{tab:graph_results}
\setlength{\tabcolsep}{6pt}
\renewcommand{\arraystretch}{1.15}
\scalebox{0.97}{
\begin{tabular}{p{0.34\textwidth}*{4}{>{\centering\arraybackslash}p{0.14\textwidth}}}
\hline
Methods & BZR & COX2 & PROTEINS & ENZYMES \\
\hline\hline
\multicolumn{5}{c}{1-shot} \\
\hline
GCN~\cite{kipf2017semi} (ICLR 2017) & 45.06 $\pm$ 16.30 & 43.84 $\pm$ 13.94 & 51.66 $\pm$ 10.87 & 19.30 $\pm$ 6.36 \\
GAT~\cite{velickovic2018gat} (ICLR 2018) & 46.28 $\pm$ 15.26 & 51.72 $\pm$ 13.70 & 51.33 $\pm$ 11.02 & 20.24 $\pm$ 6.39 \\
\hline
DGI/InfoGraph~\cite{velickovic2019dgi} (ICLR 2019) & 49.07 $\pm$ 20.34 & 53.14 $\pm$ 14.38 & 50.32 $\pm$ 13.47 & 17.73 $\pm$ 7.89 \\
GraphCL~\cite{you2020graphcl} (NeurIPS 2020) & 50.07 $\pm$ 21.38 & 50.94 $\pm$ 13.78 & 50.69 $\pm$ 10.92 & 19.73 $\pm$ 7.30 \\
\hline
GraphPrompt~\cite{liu2023graphprompt} (WWW 2023) & 52.20 $\pm$ 19.27 & 54.54 $\pm$ 16.32 & 53.61 $\pm$ 8.90 & 21.43 $\pm$ 6.60 \\
MultiGPrompt~\cite{yu2024multigprompt} (WWW 2024) & \underline{56.49 $\pm$ 19.48} & 54.63 $\pm$ 16.14 & 55.01 $\pm$ 10.19 & \underline{21.73 $\pm$ 6.54} \\
ProNoG~\cite{pronog2024nonhomophilic} (KDD 2025) & 50.43 $\pm$ 11.87 & \textbf{55.88 $\pm$ 14.30} & \underline{55.71 $\pm$ 11.10} & 21.45 $\pm$ 6.74 \\
\hline
\rowcolor{orange!10} \textbf{TPGC (Ours)} & \textbf{56.87 $\pm$ 18.50} & \underline{55.05 $\pm$ 15.09} & \textbf{55.73 $\pm$ 9.13} & \textbf{22.10 $\pm$ 6.37} \\
\hline
\multicolumn{5}{c}{5-shot} \\
\hline
GCN~\cite{kipf2017semi} (ICLR 2017) & 51.43 $\pm$ 13.32 & 47.49 $\pm$ 11.32 & 51.66 $\pm$ 10.88 & 23.47 $\pm$ 5.48 \\
GAT~\cite{velickovic2018gat} (ICLR 2018) & 52.84 $\pm$ 13.02 & 52.32 $\pm$ 10.21 & 53.93 $\pm$ 9.03 & 24.92 $\pm$ 5.30 \\
\hline
DGI/InfoGraph~\cite{velickovic2019dgi} (ICLR 2019) & 52.57 $\pm$ 18.14 & 54.52 $\pm$ 15.36 & 48.21 $\pm$ 12.35 & 21.69 $\pm$ 5.98 \\
GraphCL~\cite{you2020graphcl} (NeurIPS 2020) & 54.11 $\pm$ 16.63 & 54.09 $\pm$ 17.31 & 53.69 $\pm$ 11.92 & 21.57 $\pm$ 5.20 \\
\hline
GraphPrompt~\cite{liu2023graphprompt} (WWW 2023) & 54.60 $\pm$ 10.53 & 54.35 $\pm$ 14.78 & 54.73 $\pm$ 8.87 & 25.06 $\pm$ 7.56 \\
MultiGPrompt~\cite{yu2024multigprompt} (WWW 2024) & \underline{61.15 $\pm$ 11.63} & 56.30 $\pm$ 12.46 & 56.16 $\pm$ 9.23 & \underline{26.70 $\pm$ 6.33} \\
ProNoG~\cite{pronog2024nonhomophilic} (KDD 2025) & 59.69 $\pm$ 10.88 & \textbf{57.47 $\pm$ 11.79} & \underline{56.43 $\pm$ 9.28} & 25.36 $\pm$ 6.90 \\
\hline
\rowcolor{orange!10} \textbf{TPGC (Ours)} & \textbf{61.54 $\pm$ 11.64} & \underline{56.48 $\pm$ 13.06} & \textbf{56.69 $\pm$ 8.67} & \textbf{27.13 $\pm$ 6.44} \\
\hline
\end{tabular}
}
\end{table*}

\subsubsection{Implementation Details}

\noindent \textbf{Implementation details of baselines.} For each baseline, we implement its released official code, which is configured according to the recommended settings reported in its corresponding reference. 
For the end-to-end graph neural network baselines, GCN is implemented as a 3-layer architecture with a hidden dimension of 256. GAT is implemented as a 2-layer architecture with a hidden dimension of 64 and 8 attention heads. 
For the graph pre-training baselines, DGI adopts a 1-layer GCN backbone with a hidden dimension of 256 and uses PReLU as the activation function. GraphCL also adopts a 1-layer GCN backbone with a hidden dimension of 256, where edge dropping is used as the graph augmentation strategy with an augmentation ratio of 0.2. For the graph prompt learning baselines, GPPT uses a 2-layer GraphSAGE backbone with a hidden dimension of 256, where the mean aggregator is adopted in the GraphSAGE encoder. GraphPrompt uses a 2-layer GCN backbone with a hidden dimension of 256. MultiGPrompt uses a 1-layer GCN backbone with a hidden dimension of 256, and employs DGI, GraphCL, and LP as pre-training tasks with task weights of 0.9, 0.9, and 0.1, respectively. ProNoG uses a 1-layer GCN backbone with a default hidden dimension of 256, while the hidden dimension is set to 64 on \textit{PROTEINS} and \textit{ENZYMES}; for pre-training, DSSL~\cite{dssl2022} is used on \textit{ENZYMES}, LP is used on \textit{PROTEINS}, and GraphCL is used on the remaining datasets.

\noindent \textbf{Implementation details of our proposed TPGC.} All experiments are conducted on a single NVIDIA GeForce RTX 3090 GPU. TPGC uses a 1-layer GCN backbone with a hidden dimension of 256. In the prompt initialization stage, the number of pre-training epochs is set to 1. The adopted pre-training tasks include DGI, GraphCL, LP, and DSSL, with default task weights of 0.9, 0.9, 0.1, and 0, respectively, where DSSL is enabled only on \textit{ENZYMES}. The default top-k sampling value is set to 10. In addition, the auxiliary dataset is selected from datasets of the same type as the target dataset; for example, when the target dataset is \textit{Cora}, the auxiliary dataset is chosen as \textit{Citeseer}, since both belong to citation network datasets. To ensure a fair comparison, both the subsequent pre-training stage and the downstream task stage are conducted under the same settings as MultiGPrompt~\cite{yu2024multigprompt}.

\begin{figure*}[!htbp]
    \centering
    \begin{minipage}{\textwidth}
        \centering
        \includegraphics[width=0.75\textwidth]{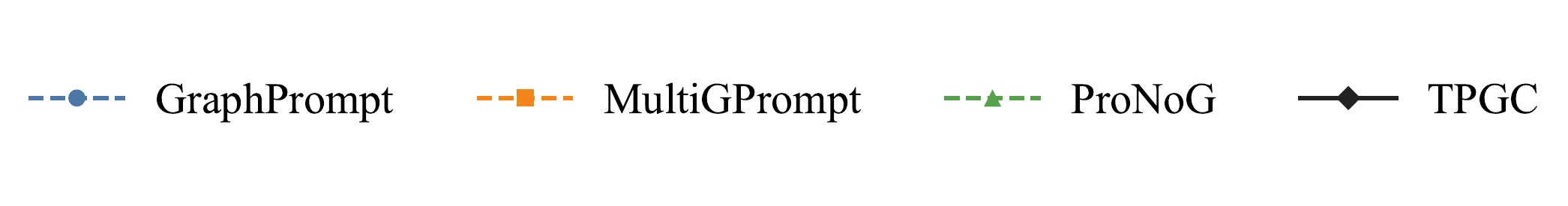}
    \end{minipage}
    \vspace{-0em}
    \begin{minipage}[t]{0.4\textwidth}
        \vspace{0pt}
        \centering
        {\small\bfseries Node Classification\par}
    \end{minipage}%
    \hspace{0.01\textwidth}%
    \begin{minipage}[t]{0.4\textwidth}
        \vspace{0pt}
        \centering
        {\small\bfseries Graph Classification\par}
    \end{minipage}
    \vspace{-0em}

    \includegraphics[width=0.8\textwidth]{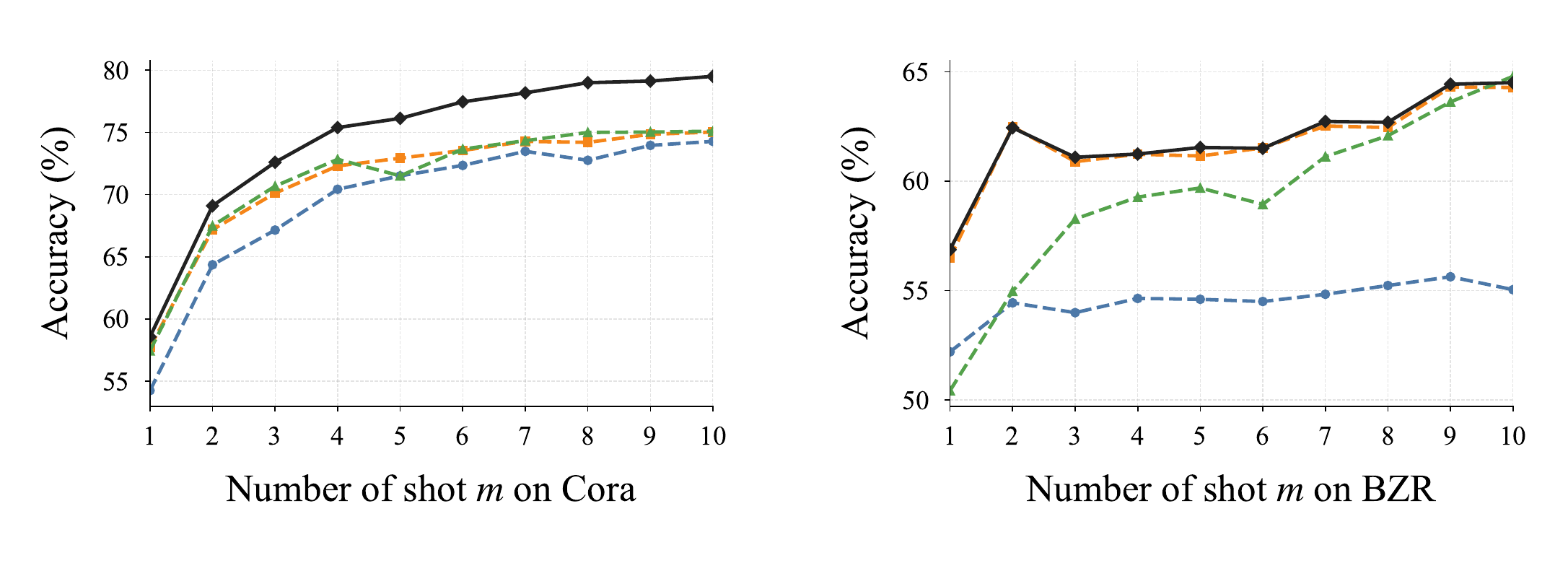}

    \vspace{-2em}

    \includegraphics[width=0.8\textwidth]{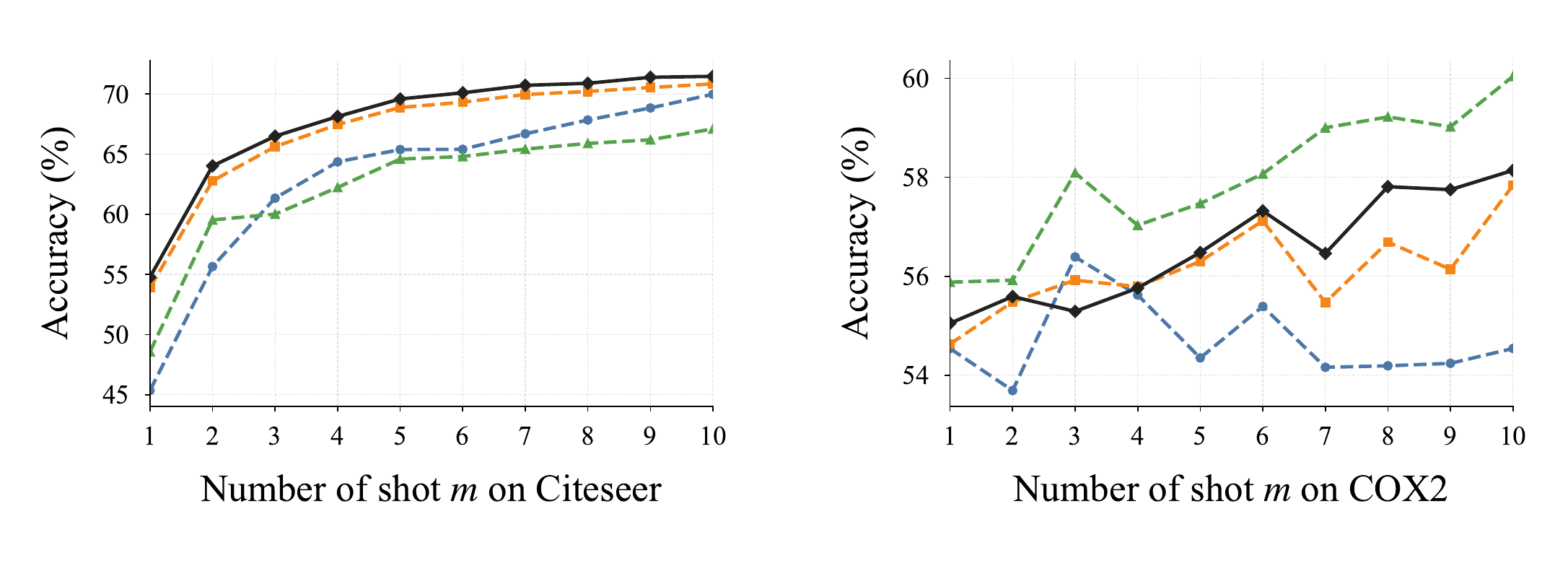}

    \vspace{-2em}

    \includegraphics[width=0.8\textwidth]{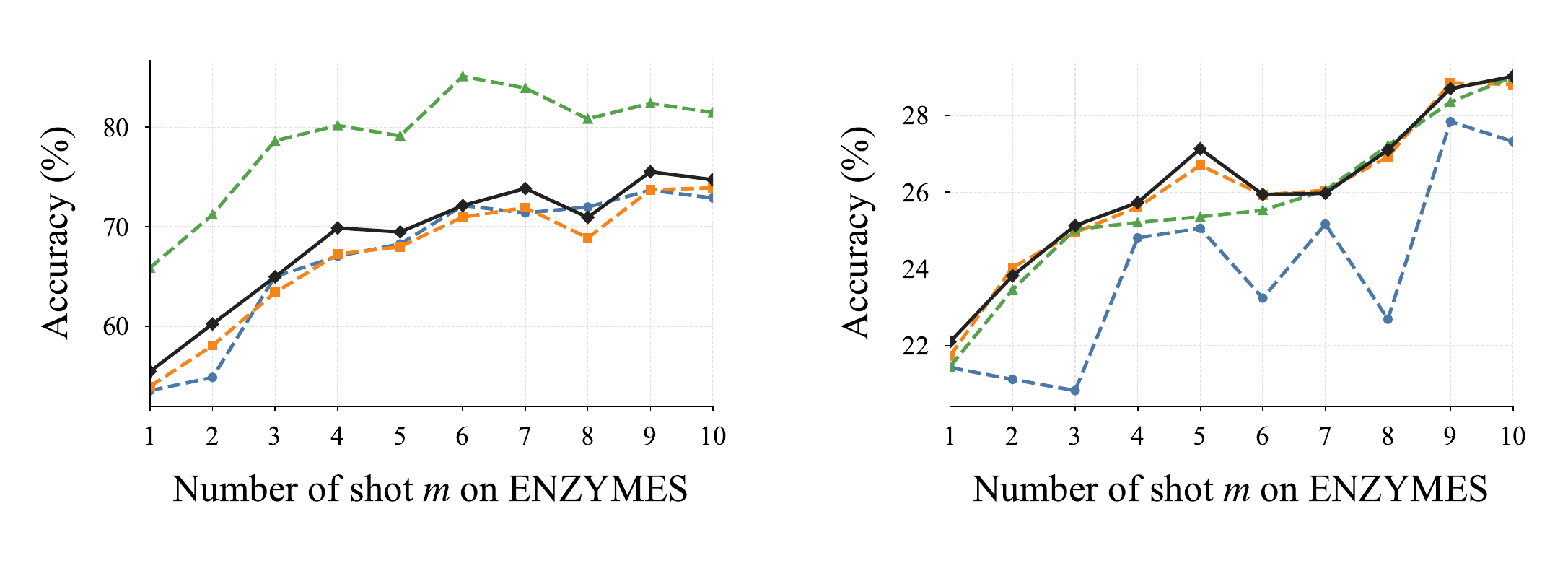}

    \vspace{-2em}

    \includegraphics[width=0.8\textwidth]{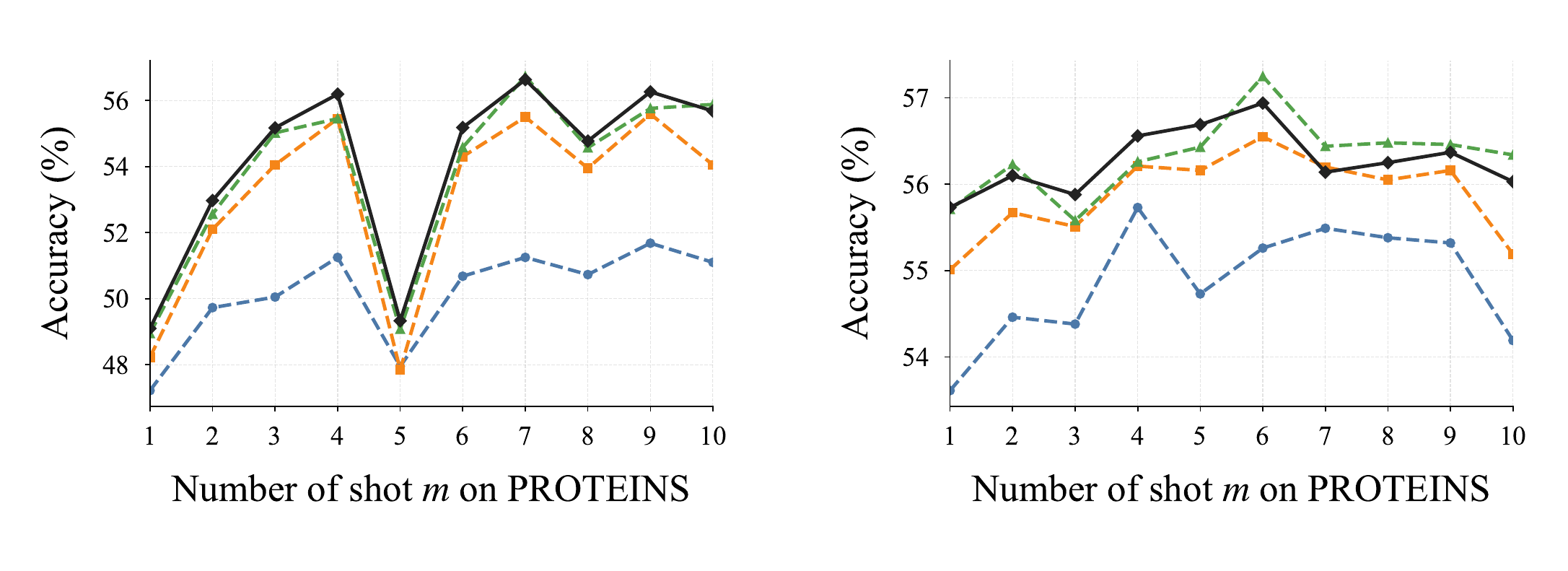}
    \vspace{-2em}
    \caption{Performance change under different shot settings on node classification and graph classification.}
    \vspace{-1em}
    \label{fig:shot_impacts}
\end{figure*}

\subsection{Performance Comparisons}

\subsubsection{Node Classification under 1-shot and 5-shot Settings} 
As shown in Table~\ref{tab:node_results}, TPGC achieves highly competitive performance across the 1-shot and 5-shot node classification settings. In particular, under each setting, TPGC obtains the best results on three out of four datasets and remains the second-best method on the remaining one. Compared with MultiGPrompt, TPGC brings consistent improvements on all four node classification datasets across both shot settings. Taking \textit{Cora} as an example, TPGC improves over MultiGPrompt by 0.84\% and 3.20\% under the 1-shot and 5-shot settings, respectively, which indicates that the proposed initialization strategy can stably enhance the effectiveness of multi-task graph prompting from extremely low-resource scenarios to relatively less sparse settings. Compared with ProNoG, TPGC maintains clear advantages on the citation-network benchmarks across both shot settings and remains competitive on \textit{PROTEINS} and \textit{ENZYMES}, although ProNoG performs better on \textit{ENZYMES}. We attribute these improvements to the collaborative effect of TPIM and SPIM: TPIM injects task-related prior into prompts before target-domain pre-training, while SPIM further guides the prompt space toward structurally informative regions through global-context-aware initialization. As a result, the initialized prompts are better aligned with downstream objectives and exhibit stronger optimization stability under limited supervision. Another important observation is that prompt-based methods, including GraphPrompt, MultiGPrompt, ProNoG, and TPGC, generally perform better than traditional graph pre-training baselines such as DGI and GraphCL across different shot settings, which further verifies that prompt-based adaptation is more effective than directly transferring pre-trained graph encoders in low-resource node classification.

\subsubsection{Graph Classification under 1-shot and 5-shot Settings} 
We further report the graph classification results under the 1-shot and 5-shot settings in Table~\ref{tab:graph_results}. The overall trend is largely consistent with that observed in node classification: TPGC remains one of the strongest methods across different supervision budgets and achieves either the best or the second-best performance on most graph classification benchmarks. More specifically, under the 1-shot and 5-shot settings, TPGC obtains the best results on \textit{BZR}, \textit{PROTEINS}, and \textit{ENZYMES}, while ranking second on \textit{COX2}. Compared with MultiGPrompt, TPGC achieves improvements on all four graph classification datasets across both shot settings. Taking \textit{PROTEINS} as an example, TPGC improves over MultiGPrompt by 0.72\% and 0.53\% under the 1-shot and 5-shot settings, respectively, which shows that the proposed prompt initialization strategy is also effective for multi-task prompted graph classification. Compared with ProNoG, these two methods show different strengths on graph classification benchmarks: TPGC is more competitive on \textit{ENZYMES} across both shot settings and also achieves stronger results on \textit{BZR} and \textit{PROTEINS} under the 1-shot and 5-shot settings, whereas ProNoG performs better on some datasets such as \textit{COX2}. Overall, these results suggest that TPGC also delivers strong and reliable performance on graph classification tasks.

\begin{table*}[t]
\centering
\normalsize
\caption{Comparison of downstream-stage runtime and tunable parameter count on \textit{ENZYMES} under the 1-shot setting with 50 downstream training epochs. Runtime is reported in milliseconds (ms).}
\label{tab:complexity_runtime}
\setlength{\tabcolsep}{10pt}
\renewcommand{\arraystretch}{1.15}
\scalebox{0.87}{
\begin{tabular}{lcc}
\hline
Methods & Total Runtime (ms) & Tunable Parameters \\
\hline\hline
MultiGPrompt~\cite{yu2024multigprompt} (WWW 2024) & 2859.65 & 522 \\
ProNoG~\cite{pronog2024nonhomophilic} (KDD 2025) & 5358.24 & 2564 \\ \hline
\rowcolor{orange!10} \textbf{TPGC (Ours)} & 2885.47 & 522 \\
\hline
\end{tabular}
}
\end{table*}

\begin{figure*}[t]
    \centering
    \includegraphics[width=0.88\textwidth]{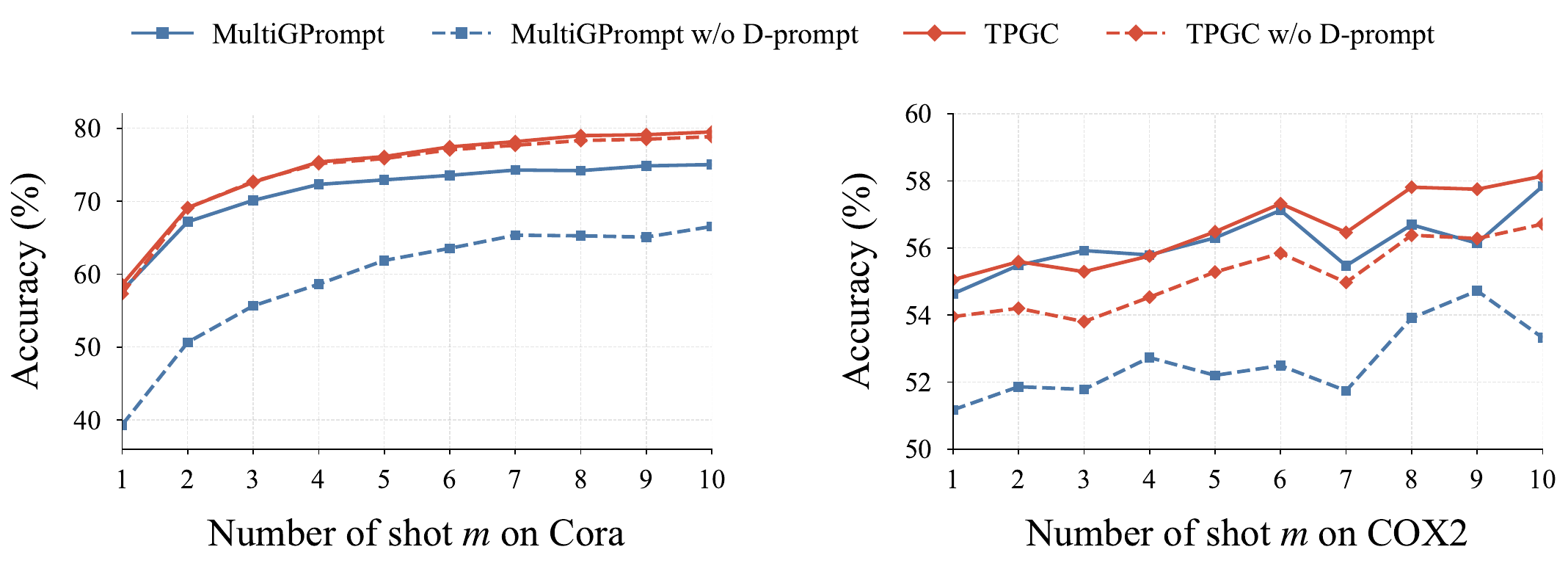}
    \caption{Performance comparison before and after removing downstream-task-specific prompts (abbreviated as D-prompt) on \textit{Cora} node classification and \textit{COX2} graph classification. Solid lines denote the original methods, while dashed lines denote the variants that discard D-prompt and only use pre-trained prompt vectors for downstream classification.}
    \label{fig:remove_downstream_prompt_effect}
\end{figure*}

\subsubsection{Strong Robustness under Different Shots} 
To further evaluate the robustness of TPGC under different few-shot settings, we vary the number of shots $m$ from 1 to 10 for both node classification and graph classification tasks. The corresponding results are illustrated in Figure~\ref{fig:shot_impacts}, and we make several observations below. 
\begin{itemize}
    \item First, TPGC consistently achieves competitive performance across different shot settings and remains one of the strongest methods on most datasets. In particular, under low-shot settings (e.g., $m \leq 5$), TPGC generally performs better than or on par with strong prompt-learning baselines such as GraphPrompt, MultiGPrompt, and ProNoG, demonstrating its effectiveness when only very limited labeled data are available.
    \item Second, as the number of shots increases, the performance of all methods generally improves, which is expected since more supervision is provided. Nevertheless, TPGC still maintains clear advantages or highly competitive results in most cases, indicating that the proposed prompt initialization strategy is not only beneficial in extremely low-resource scenarios but also remains effective when more labeled samples are given. 
\end{itemize}

\begin{table*}[t]
\centering
\normalsize
\caption{Effectiveness of key component in prompt initialization on node classification datasets.}
\label{tab:ablation_node_results}
\setlength{\tabcolsep}{6pt}
\renewcommand{\arraystretch}{1.15}
\scalebox{0.9}{
\begin{tabular}{p{0.24\textwidth}*{2}{>{\centering\arraybackslash}p{0.08\textwidth}}*{4}{>{\centering\arraybackslash}p{0.13\textwidth}}}
\hline
Methods & SPIM & TPIM & Cora & Citeseer & PROTEINS & ENZYMES \\
\hline\hline
\multicolumn{7}{c}{1-shot} \\
\hline
Random Init. &  &  & 57.73 & 53.89 & 48.23 & 53.95 \\
TPGC w/o T & $\surd$ &  & 57.95 & 53.73 & 46.92 & 53.54 \\
TPGC w/o S &  & $\surd$ & 57.09 & 53.28 & 47.30 & 53.90 \\ \hline
\rowcolor{orange!10} \textbf{TPGC (Ours)} & $\surd$ & $\surd$ & \textbf{58.57} & \textbf{54.77} & \textbf{49.10} & \textbf{55.45} \\
\hline
\multicolumn{7}{c}{5-shot} \\
\hline
Random Init. &  &  & 72.93 & 68.88 & 47.85 & 67.99 \\
TPGC w/o T & $\surd$ &  & 72.83 & 68.90 & 48.67 & 66.81 \\
TPGC w/o S &  & $\surd$ & 72.67 & 68.57 & 48.21 & 66.81 \\ \hline
\rowcolor{orange!10} \textbf{TPGC (Ours)} & $\surd$ & $\surd$ & \textbf{76.13} & \textbf{69.09} & \textbf{49.23} & \textbf{69.47} \\
\hline
\end{tabular}
}
\end{table*}

\begin{table*}[t]
\centering
\normalsize
\caption{Effectiveness of key component in prompt initialization on graph classification datasets.}
\label{tab:ablation_graph_results}
\setlength{\tabcolsep}{6pt}
\renewcommand{\arraystretch}{1.15}
\scalebox{0.9}{
\begin{tabular}{p{0.24\textwidth}*{2}{>{\centering\arraybackslash}p{0.08\textwidth}}*{4}{>{\centering\arraybackslash}p{0.13\textwidth}}}

\hline
Methods & SPIM & TPIM & BZR & COX2 & PROTEINS & ENZYMES \\
\hline\hline
\multicolumn{7}{c}{1-shot} \\
\hline
Random Init. &  &  & 56.49 & 54.63 & 55.11 & 21.73 \\
TPGC w/o T & $\surd$ &  & 56.18 & 53.67 & 54.56 & 21.53 \\
TPGC w/o S &  & $\surd$ & 55.17 & 53.15 & 54.13 & 21.03 \\ \hline
\rowcolor{orange!10} \textbf{TPGC (Ours)} & $\surd$ & $\surd$ & \textbf{56.87} & \textbf{55.05} & \textbf{55.73} & \textbf{22.10} \\
\hline
\multicolumn{7}{c}{5-shot} \\
\hline
Random Init. &  &  & 61.15 & 56.30 & 56.16 & 26.70 \\
TPGC w/o T & $\surd$ &  & 60.57 & 54.48 & 55.49 & 25.90 \\
TPGC w/o S &  & $\surd$ & 60.20 & 54.18 & 55.08 & 25.70 \\ \hline
\rowcolor{orange!10} \textbf{TPGC (Ours)} & $\surd$ & $\surd$ & \textbf{61.54} & \textbf{56.48} & \textbf{56.69} & \textbf{27.13} \\
\hline
\end{tabular}
}
\end{table*}

Overall, these results further verify the robustness and generalization of TPGC across both node-level and graph-level tasks under a wide range of few-shot learning settings.

\subsubsection{Downstream Complexity Analysis} 
In addition to accuracy comparisons, we further conduct a downstream-stage complexity study on \textit{MultiGPrompt}, \textit{ProNoG}, and our \textit{TPGC}. The overall runtime comparison and tunable-parameter comparison are reported in Table~\ref{tab:complexity_runtime}. Since TPGC follows the same downstream implementation as MultiGPrompt, both methods have exactly the same number of tunable parameters in the downstream stage, namely 522, and their computational overheads also remain very close. In contrast, ProNoG introduces a much larger downstream parameter budget of 2564 and substantially heavier computation during downstream inference. Specifically, ProNoG requires additional neighbor retrieval, prompt-based transformation, similarity-weighted aggregation, and meta-network inference for each test node before prototype matching. Such a multi-step neighbor-aware pipeline introduces substantially higher computational overhead than TPGC.
Taking \textit{ENZYMES} as an example, the total downstream runtime of ProNoG reaches 5358.24 ms, which is about 1.86$\times$ that of TPGC (2885.47 ms). This result suggests that the relatively strong accuracy of ProNoG on \textit{ENZYMES} is associated with both a larger downstream parameter scale and a noticeably heavier computational overhead, whereas TPGC maintains a more favorable efficiency--effectiveness trade-off while still achieving highly competitive predictive performance.

\begin{figure*}[t]
    \centering
    \includegraphics[width=0.88\linewidth]{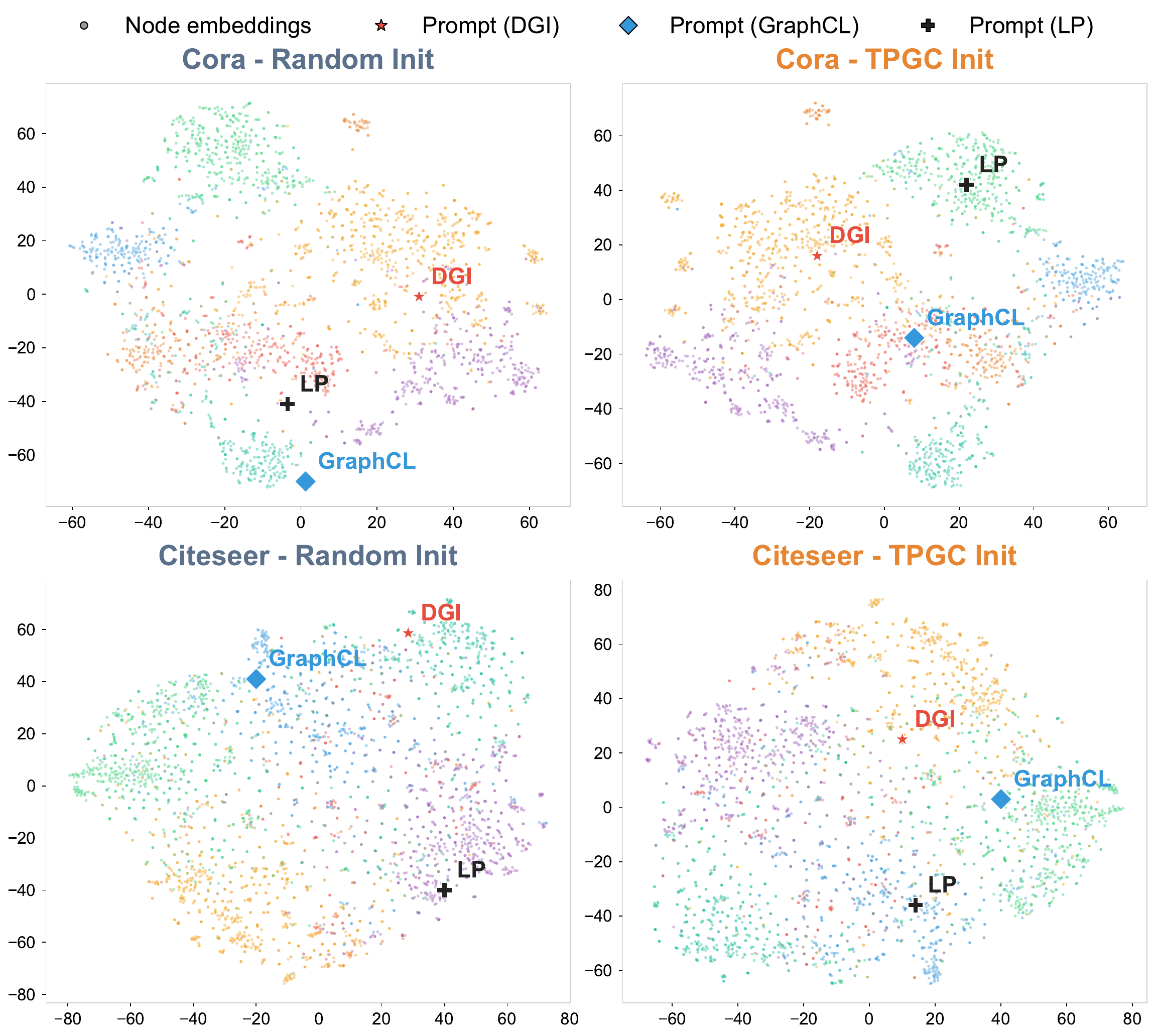}
    \caption{Visualization (t-SNE) of initialized prompt vectors and graph node embeddings. DGI, GraphCL, and LP correspond to the three default pretext tasks adopted in the pre-training stage, and the plotted prompt vectors illustrate how task-specific prompt initialization is positioned in the graph representation space.}
    \label{tsne_visualization}
\end{figure*}

\subsection{Ablation Studies}

\subsubsection{Discussion on the Pivotal Role of Prompt Initialization} 
To further compare the behavior of multi-task pre-trained prompt learning under different prompt initialization strategies with the most related work MultiGPrompt, we conduct an ablation study by removing the downstream-task-specific prompt and only keeping the initialized prompt. In this setting, both MultiGPrompt and TPGC discard the prompt introduced specifically for downstream adaptation, and only the pre-trained prompt vectors are used to guide downstream classification.
As shown in Figure~\ref{fig:remove_downstream_prompt_effect}, MultiGPrompt relies heavily on the downstream-task-specific prompt. After removing it, the accuracy drops substantially in all cases, typically by around 5 percentage points, and the degradation on \textit{Cora} can reach roughly 10--20 percentage points across different shot settings. This observation suggests that randomly initialized pre-training prompts still contain high information entropy and strong optimization uncertainty, so the learned prompt vectors alone are not sufficiently stable to guide downstream classification without additional task-specific adaptation. In contrast, TPGC effectively alleviates this issue through a more informative initialization strategy. After discarding the downstream-task-specific prompt, TPGC is much less sensitive to this removal, with a slight performance drop. This result indicates that, by jointly injecting task priors and global contextual information into the prompt initialization process, TPGC learns pre-trained prompt vectors that are better aligned with downstream decision boundaries and can provide more stable, transferable, and task-aware guidance for downstream classification.

\subsubsection{Effectiveness of Key Component in Prompt Initialization} 
To verify the effectiveness of each component (i.e., TPIM and SPIM) in our proposed prompt initialization strategy, we further conduct ablation studies under the 1-shot and 5-shot settings and compare the full model with three representative variants. The corresponding results are reported in Table~\ref{tab:ablation_node_results} and Table~\ref{tab:ablation_graph_results}. Overall, the full TPGC achieves the strongest or tied-strongest performance on almost all datasets across both shot settings, which confirms the effectiveness of combining TPIM and SPIM in prompt initialization. 
Specifically, \textit{Random Init.} denotes the original MultiGPrompt-style random initialization without either TPIM or SPIM. \textit{TPGC w/o T} denotes TPGC without TPIM, which directly uses the prompt vectors obtained after structural aggregation on the auxiliary graph as initialization for target-domain pre-training, without performing auxiliary-graph pre-training in advance. \textit{TPGC w/o S} denotes TPGC without SPIM, where the top-k similarity-based sampling operation is removed and replaced with global average pooling over all nodes. The full \textit{TPGC} jointly incorporates both TPIM and SPIM, so that prompt initialization can simultaneously encode task-related prior and transferable structural information. From Table~\ref{tab:ablation_node_results} and Table~\ref{tab:ablation_graph_results}, we can draw the following observations.

\begin{itemize}
    \item First, removing either TPIM or SPIM generally leads to performance degradation under both the 1-shot and 5-shot settings, which indicates that both components are beneficial. In particular, the drop of \textit{TPGC w/o T} verifies that auxiliary-graph pre-training is important for injecting task-aware prior before target-domain optimization, while the weaker performance of \textit{TPGC w/o S} on several datasets shows that replacing top-k sampling with simple global averaging makes it harder to preserve informative structural patterns. 
    \item Second, compared with \textit{Random Init.}, the full TPGC yields more consistent gains across both node classification and graph classification tasks in the two shot settings, suggesting that the proposed prompt initialization is more reliable than purely random initialization. 
\end{itemize}

In summary, these results demonstrate that TPIM and SPIM are complementary, and their combination provides the most effective prompt initialization strategy for multi-task graph pre-training under different few-shot settings.

\begin{figure*}[t]
    \centering
    \includegraphics[width=0.88\linewidth]{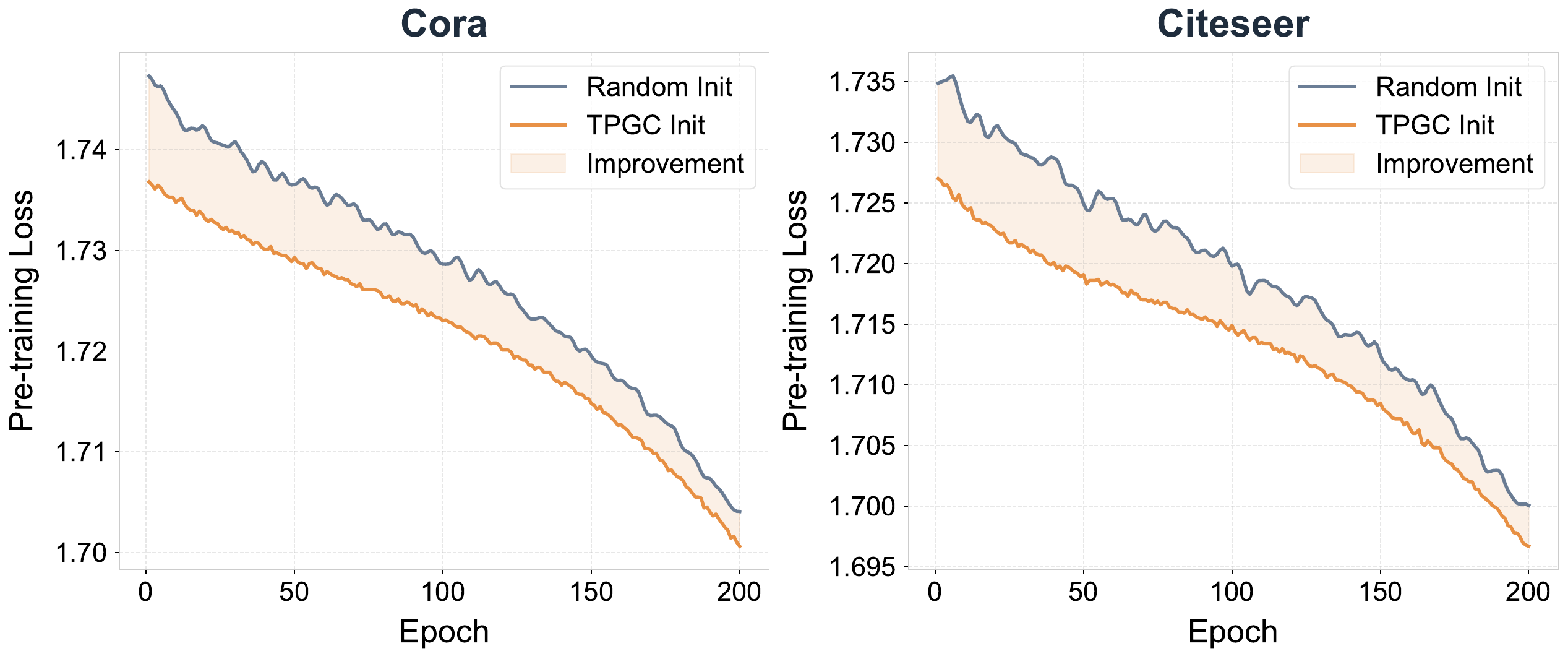}
    \caption{Visualization of pre-training loss convergence under random initialization and TPGC initialization.}
    \label{loss_convergence}
\end{figure*}

\subsection{Visualization}

\subsubsection{Visualization of Prompt Vectors} 
To further understand the effect of prompt initialization from a representation-space perspective, we visualize the initialized prompt vectors together with graph node embeddings after t-SNE dimensionality reduction, as shown in Figure~\ref{tsne_visualization}. Overall, the visualization clearly shows that TPGC produces prompt vectors that are better aligned with the structure of the graph representation space, whereas random initialization tends to place prompts in less informative regions. This observation provides intuitive evidence for why the proposed prompt initialization strategy improves subsequent multi-task graph pre-training and downstream adaptation.
As shown in the left of Figure~\ref{tsne_visualization}, under random initialization, the prompt vectors are scattered in arbitrary noisy regions of the representation space and are often far away from meaningful node clusters. Such a distribution indicates that the initialized prompts are poorly aligned with the intrinsic organization of graph embeddings. As a result, the prompts cannot provide informative guidance at the beginning of pre-training, and the optimization process must spend additional effort to move them toward task-relevant regions.
In contrast, as shown in the right of Figure~\ref{tsne_visualization}, with TPGC initialization, the prompt vectors are located much closer to high-density and semantically meaningful regions of the node embedding distribution. Instead of lying in isolated noisy areas, they are naturally aligned with the overall geometry of the representation space and better match the underlying node clusters. This phenomenon directly supports the role of SPIM, namely, aligning prompt initialization with high-information regions in the graph representation space by injecting transferable structural prior from the auxiliary graph.

\subsubsection{Visualization of Convergence}
In addition to the representation-space visualization, the loss convergence curves in Figure~\ref{loss_convergence} further confirm the optimization advantage of the proposed initialization strategy. After applying TPGC-based prompt initialization, the pre-training loss decreases more rapidly and reaches a lower level in earlier stages than random initialization. This indicates that a better-aligned prompt initialization not only places the prompts in more informative regions of the representation space, but also provides a more favorable starting point for optimization. As a result, the model can enter an effective training regime more quickly, which demonstrates that TPGC helps the pre-training process converge faster and more stably.


\section{Conclusion and Future Works}
\label{sec:conclusion}
In this paper, we propose a new prompt initialization method (TPGC) for multi-task graph pre-training. The core idea of TPGC is to initialize graph prompts with both task-aware prior knowledge and transferable structural context, so that the subsequent pre-training and downstream adaptation stages can start from a more informative and stable prompt space. Extensive benchmark experiments demonstrate that the two designed modules, SPIM and TPIM, can effectively enhance prompt quality, improve few-shot performance, and promote more stable optimization across both node classification and graph classification tasks. Despite these encouraging results, we also recognize that the current design may still be limited when handling heterophilous graphs, where connected nodes often have different labels or semantics, and we leave more effective prompt initialization for such scenarios as our future works.

\section*{CRediT Authorship Contribution Statement}
\textbf{Zhiyang Qiu:} Conceptualization, Data curation, Investigation, Validation, Writing - original draft. \textbf{Yangtao Wang:} Conceptualization, Investigation, Methodology, Resources,  Writing - review \& editing, Supervision.  \textbf{Xiaocui Li:} Conceptualization, Writing - review \& editing. \textbf{Yanzhao Xie:} Conceptualization, Writing - review \& editing. \textbf{Siyuan Chen:} Conceptualization, Writing - review \& editing. \textbf{Wensheng Zhang:} Conceptualization, Supervision.

\section*{Declaration of Competing Interest}
The authors declare that they have no known competing financial interests or personal relationships that could have appeared to influence the work reported in this paper.

\section*{Data Availability}
Data are available at \textcolor{blue}{\url{https://github.com/Virgilqiu/TPGC}}.

\section*{Acknowledgments}
This work is supported by National Natural Science Foundation of China (No. 62406082, No. 62506085, No. 62394334), Guangdong Basic and Applied Basic Research Foundation (No. 2023A1515110650, No. 2023A1515110659), Guangzhou Science and Technology Planning Project (No. 2024A03J0013, No. 2025A04J4590), Guangdong Provincial Department of Education Innovation Strong School Youth Innovation Talent Project (No. 2023KQNCX055), and Youth S\&T Talent
Support Programme of Guangdong Provincial Association for Science and Technology (GDSTA) (No. SKXRC2026111).

\bibliographystyle{cas-model2-names}

\bibliography{References}



\end{document}